\documentclass[11pt,a4paper]{article}
\usepackage[hyperref]{acl2020}
\usepackage{times}
\usepackage{tikz}
\usepackage{subfigure}
\usepackage{xcolor}
\usepackage{tcolorbox}
\usepackage{latexsym}
\usepackage{graphicx}
\usepackage{pgfplots}
\usepackage{amsmath,amssymb}
\usepackage{bbm}
\usepackage{booktabs}
\usepackage{multirow}
\usepackage{amsmath}
\usepackage{amssymb}
\usepackage{xspace}

\usepackage{microtype}
\usetikzlibrary{mindmap,backgrounds,fit}
\usetikzlibrary{shadows}
\aclfinalcopy % Uncomment this line for the final submission
\definecolor{ublue}{rgb}{0.152,0.250,0.545}
\definecolor{ugreen}{rgb}{0.0,0.5,0}

\usepackage{amssymb}% http://ctan.org/pkg/amssymb
\usepackage{pifont}% http://ctan.org/pkg/pifont
\newcommand{\cmark}{\ding{51}}%

\title{Learning Architectures from an Extended Search Space \\ for Language Modeling}

\author{
    Yinqiao Li$^1$,
    Chi Hu$^1$,
    Yuhao Zhang$^1$,
    Nuo Xu$^1$,
    Yufan Jiang$^1$, \\
    \textbf{Tong Xiao$^{1,2}$\thanks{\xspace\xspace Corresponding author.}},
    \textbf{Jingbo Zhu$^{1,2}$},
    \textbf{Tongran Liu$^3$},
    \textbf{Changliang Li$^4$} \\
    $^{1}$NLP Lab, Northeastern University, Shenyang, China\\
    $^{2}$NiuTrans Research, Shenyang, China \\
    $^{3}$CAS Key Laboratory of Behavioral Science, Institute of Psychology, CAS, Beijing, China \\
    $^{4}$Kingsoft AI Lab, Beijing, China \\
    {\tt
        li.yin.qiao.2012@hotmail.com, 
        }\\
    {\tt
        \{huchinlp,yoohao.zhang\}@gmail.com, 
        }\\
    {\tt
        \{xunuo0629,jiangyufan2018\}@outlook.com,
        } \\
    {\tt
        \{xiaotong,zhujingbo\}@mail.neu.edu.com,
    } \\
    {\tt
        liutr@psych.ac.cn,lichangliang@kingsoft.com
    } \\
}

\date{}

\begin{document}
\maketitle
\begin{abstract}

Neural architecture search (NAS) has advanced significantly in recent years but most NAS systems restrict search to learning architectures of a recurrent or convolutional cell. In this paper, we extend the search space of NAS. In particular, we present a general approach to learn both intra-cell and inter-cell architectures (call it ESS). For a better search result, we design a joint learning method to perform intra-cell and inter-cell NAS simultaneously. We implement our model in a differentiable architecture search system. For recurrent neural language modeling, it outperforms a strong baseline significantly on the PTB and WikiText data, with a new state-of-the-art on PTB. Moreover, the learned architectures show good transferability to other systems. E.g., they improve state-of-the-art systems on the CoNLL and WNUT named entity recognition (NER) tasks and CoNLL chunking task, indicating a promising line of research on large-scale pre-learned architectures.

\end{abstract}

\section{Introduction}

Neural models have shown remarkable performance improvements in a wide range of natural language processing (NLP) tasks. Systems of this kind can broadly be characterized as following a neural network design: we model the problem via a pre-defined neural architecture, and the resulting network is treated as a black-box family of functions for which we find parameters that can generalize well on test data. This paradigm leads to many successful NLP systems based on well-designed architectures. The earliest of these makes use of recurrent neural networks (RNNs) for representation learning  \cite{bahdanau2015jointlearning,wu2016google}, whereas recent systems have successfully incorporated fully attentive models into language generation and understanding \cite{vaswani2017attention}.

In designing such models, careful engineering of the architecture plays a key role for the state-of-the-art though it is in general extremely difficult to find a good network structure. The next obvious step is toward automatic architecture design. A popular method to do this is neural architecture search (NAS). In NAS, the common practice is that we first define a search space of neural networks, and then find the most promising candidate in the space by some criteria. Previous efforts to make NAS more accurate have focused on improving search and network evaluation algorithms. But the search space is still restricted to a particular scope of neural networks. For example, most NAS methods are applied to learn the topology in a recurrent or convolutional cell, but the connections between cells are still made in a heuristic manner as usual \cite{zoph2017reinforcement,elsken2019efficient}.

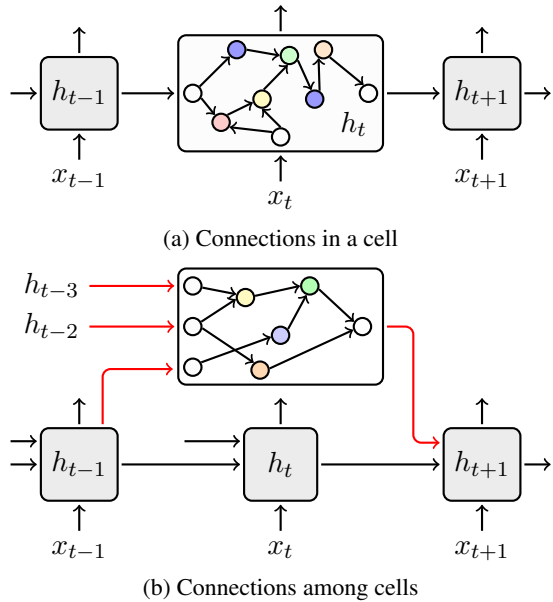
\begin{figure}[t!]
  \centering
  \tikzstyle{rnnnode1} = [rounded corners=3pt,inner sep=4pt,minimum height=2.5em,minimum width=2.5em,draw,thick,fill=lightgray!30]
  \tikzstyle{rnnnode2} = [rounded corners=3pt,inner sep=4pt,minimum height=4em,minimum width=7em,draw,thick,fill=lightgray!5]
  \tikzstyle{cellnode} = [minimum size=0.3em,inner sep=2.3pt,circle,draw,thick,fill=lightgray!5]
  \tikzstyle{standard} = [rounded corners=3pt,thick]
  \centering
    \begin{tikzpicture}
        \node [rnnnode1,anchor=west] (h1) at (0,0) {$h_{t-1}$};
        \node [anchor=north] (x1) at ([yshift=-1em]h1.south) {$x_{t-1}$};
        \node [rnnnode2,anchor=west] (h2) at ([xshift=2em]h1.east) {};
        \node [anchor=north] (x2) at ([yshift=-1em]h2.south) {$x_{t}$};
        \node [rnnnode1,anchor=west] (h3) at ([xshift=2em]h2.east) {$h_{t+1}$};
        \node [anchor=north] (x3) at ([yshift=-1em]h3.south) {$x_{t+1}$};
        \node [cellnode,anchor=west] (cell1) at ([xshift=0.2em]h2.west) {};
        \node [cellnode,anchor=south] (cell2) at ([yshift=0.2em]h2.south) {};
        \node [cellnode,anchor=north,fill=blue!40] (cell3) at ([xshift=-1.5em,yshift=-0.2em]h2.north) {};
        \node [cellnode,anchor=east] (cell4) at ([xshift=-0.2em]h2.east) {};
        \node [cellnode,anchor=west,fill=red!20] (cell5) at ([xshift=0.3em,yshift=-1em]cell1.east) {};
        \node [cellnode,anchor=west,fill=yellow!30] (cell6) at ([xshift=1.7em,yshift=-0.2em]cell1.east) {};
        \node [cellnode,anchor=north,fill=green!20] (cell7) at ([xshift=0.3em,yshift=-0.4em]h2.north) {};
        \node [cellnode,anchor=west,fill=blue!40] (cell8) at ([xshift=3.5em,yshift=-0.2em]cell1.east) {};
        \node [cellnode,anchor=west,fill=orange!20] (cell9) at ([xshift=3.8em,yshift=1.5em]cell1.east) {};

        \node [anchor=south] (label) at ([xshift=2.5em,yshift=0.2em]h2.south) {$h_{t}$};
        \node [anchor=north] (a) at ([yshift=-2.3em]h2.south) {\small (a) Connections in a cell };

        %bottom
        \node [rnnnode2,anchor=north,fill=white] (h7) at ([yshift=-6em]h2.center) {};
        \node [rnnnode1,anchor=north] (h4) at ([yshift=-11.5em]h1.center) {$h_{t-1}$};
        \node [anchor=north] (x4) at ([yshift=-1em]h4.south) {$x_{t-1}$};
        \node [rnnnode1,anchor=north] (h5) at ([yshift=-11.5em]h2.center) {$h_{t}$};
        \node [anchor=north] (x5) at ([yshift=-1em]h5.south) {$x_{t}$};
        \node [rnnnode1,anchor=north] (h6) at ([yshift=-11.5em]h3.center) {$h_{t+1}$};
        \node [anchor=north] (x6) at ([yshift=-1em]h6.south) {$x_{t+1}$};

        \node [cellnode,anchor=west] (intra1) at ([xshift=0.2em]h7.west) {};
        \node [cellnode,anchor=west] (intra2) at ([xshift=0.2em,yshift=1.4em]h7.west) {};
        \node [cellnode,anchor=west] (intra3) at ([xshift=0.2em,yshift=-1.4em]h7.west) {};
        \node [cellnode,anchor=west,fill=yellow!30] (intra4) at ([xshift=2em,yshift=1em]h7.west) {};
        \node [cellnode,anchor=west,fill=blue!20] (intra5) at ([xshift=3.2em,yshift=-0.3em]h7.west) {};
        \node [cellnode,anchor=west,fill=orange!30] (intra6) at ([xshift=2.5em,yshift=-1.5em]h7.west) {};
        \node [cellnode,anchor=west,fill=green!30] (intra7) at ([xshift=4.2em,yshift=1.4em]h7.west) {};
        \node [cellnode,anchor=west,fill=white] (intra8) at ([xshift=6em]h7.west) {};

        %\node [anchor=south] (labe2) at ([xshift=2.5em,yshift=0.2em]h7.south) {$\hat{h}$};
        \node [anchor=east] (labe3) at ([xshift=-3em,yshift=1.4em]h7.west) {$h_{t-3}$};
        \node [anchor=east] (labe4) at ([xshift=-3em]h7.west) {$h_{t-2}$};

        \node [anchor=north] (b) at ([yshift=-2.3em]h5.south) {\small (b) Connections among cells };
        %connection(up)
        \foreach \x in {1,2,3}{
            \draw [->,thick] ([yshift=0.1em]h\x.north) -- ([yshift=1em]h\x.north);
            \draw [->,thick] ([yshift=-1em]h\x.south) -- ([yshift=-0.1em]h\x.south);
        }
        \draw [->,thick] ([xshift=-1em]h1.west) -- ([xshift=-0.1em]h1.west);
        \draw [->,thick] ([xshift=0.1em]h1.east) -- ([xshift=-0.1em]h2.west);
        \draw [->,thick] ([xshift=0.1em]h3.east) -- ([xshift=1em]h3.east);
        \draw [->,thick] ([xshift=0.1em]h2.east) -- ([xshift=-0.1em]h3.west);

        \draw [->,thick] (cell1.45) -- (cell3.225);
        \draw [->,thick] (cell1.330) -- (cell5.130);
        \draw [->,thick] (cell2.150) -- (cell5.330);
        \draw [->,thick] (cell2.90) -- (cell6.270);
        \draw [->,thick] (cell5.50) -- (cell6.200);
        \draw [->,thick] (cell3.0) -- (cell7.180);
        \draw [->,thick] (cell6.80) -- (cell7.260);
        \draw [->,thick] (cell7.330) -- (cell8.110);
        \draw [->,thick] (cell8.60) -- (cell9.240);
        \draw [->,thick] (cell9.310) -- (cell4.120);

        %connection(bottom)
        \foreach \x in {4,5,6}{
            \draw [->,thick] ([yshift=0.1em]h\x.north) -- ([yshift=1em]h\x.north);
            \draw [->,thick] ([yshift=-1em]h\x.south) -- ([yshift=-0.1em]h\x.south);
        }
        \draw [->,thick] ([xshift=-1em]h4.west) -- ([xshift=-0.1em]h4.west);
        \draw [->,thick] ([xshift=0.1em]h4.east) -- ([xshift=-0.1em]h5.west);
        \draw [->,thick] ([xshift=0.1em]h5.east) -- ([xshift=-0.1em]h6.west);
        \draw [->,thick] ([xshift=0.1em]h6.east) -- ([xshift=1em]h6.east);

        \draw [->,thick] ([xshift=-1em,yshift=0.8em]h4.west) -- ([xshift=-0.1em,yshift=0.8em]h4.west);
        \draw [->,thick] ([xshift=-2em,yshift=0.8em]h5.west) -- ([xshift=-0.1em,yshift=0.8em]h5.west);

        \draw [->,thick] (intra1.30) -- (intra4.190);
        \draw [->,thick] (intra1.350) -- (intra6.130);
        \draw [->,thick] (intra2.350) -- (intra4.160);
        \draw [->,thick] (intra3.30) -- (intra5.190);
        \draw [->,thick] (intra4.10) -- (intra7.180);
        \draw [->,thick] (intra5.40) -- (intra7.250);
        \draw [->,thick] (intra6.10) -- (intra8.200);
        \draw [->,thick] (intra7.340) -- (intra8.150);

        \draw [->,standard,red] ([xshift=0.1em]h7.east) -- ([xshift=1em]h7.east) -- ([xshift=1em,yshift=-4.0em]h7.east) -- ([xshift=1.9em,yshift=-4.0em]h7.east);
        \draw [->,standard,red] ([xshift=0.8em,yshift=0.1em]h4.north) -- ([xshift=0.8em,yshift=2em]h4.north) -- ([xshift=3.2em,yshift=2em]h4.north);
        \draw [->,standard,red] ([xshift=-3em]h7.west) -- ([xshift=-0.1em]h7.west);
        \draw [->,standard,red] ([xshift=-3em,yshift=1.4em]h7.west) -- ([xshift=-0.1em,yshift=1.4em]h7.west);

    \end{tikzpicture}

  \caption{Examples of intra and inter-cell architectures. } \label{fig:inter-intra-conections}
  \label{overall}
\end{figure}

Note that the organization of these sub-networks remains important as to the nature of architecture design. For example, the first-order connectivity of cells is essential to capture the recurrent dynamics in RNNs. More recently, it has been found that additional connections of RNN cells improve LSTM models by accessing longer history on language modeling tasks \cite{melis2019mogrifier}. Similar results appear in Transformer systems. Dense connections of distant layers help in learning a deep Transformer encoder for machine translation \cite{shen2018dense}. A natural question that arises is: can we learn the connectivity of sub-networks for better architecture design?

In this paper, we address this issue by enlarging the scope of NAS and learning connections among sub-networks that are designed in either a handcrafted or automatic way (Figure \ref{fig:inter-intra-conections}). We call this the Extended Search Space method for NAS (or ESS for short). Here, we choose differentiable architecture search as the basis of this work because it is efficient and gradient-friendly. We present a general model of differentiable architecture search to handle arbitrary search space of NAS, which offers a unified framework of describing intra-cell NAS and inter-cell NAS. Also, we develop a joint approach to learning both high-level and low-level connections simultaneously. This enables the interaction between intra-cell NAS and inter-cell NAS, and thus the ability of learning the full architecture of a neural network.

Our ESS method is simple for implementation. We experiment with it in an RNN-based system for language modeling. On the PTB and WikiText data, it outperforms a strong baseline significantly by 4.5 and 2.4 perplexity scores. Moreover, we test the transferability of the learned architecture on other tasks. Again, it shows promising improvements on both NER and chunking benchmarks, and yields new state-of-the-art results on NER tasks. This indicates a promising line of research on large-scale pre-learned architectures.  More interestingly, it is observed that the inter-cell NAS is helpful in modeling rare words. For example, it yields a bigger improvement on the rare entity recognition task (WNUT) than that on the standard NER task (CoNLL).

\section{Related work}

NAS is a promising method toward AutoML \cite{automl_book}, and has been recently applied to NLP tasks \cite{david2019evolved,jiang2019idarts,liam2019random}. Several research teams have investigated search strategies for NAS. The very early approaches adopted evolutionary algorithms to model the problem \cite{Peter1994evolutionary,stanley2002evolving}, while Bayesian and reinforcement learning methods made big progresses in computer vision and NLP later \cite{bergstra2013hyperparameter,bowen2017reinforcement,zoph2017reinforcement}. More recently, gradient-based methods were successfully applied to language modeling and image classification based on RNNs and CNNs \cite{liu2019darts}. In particular, differentiable architecture search has been of great interest to the community because of its efficiency and compatibility to off-the-shelf tools of gradient-based optimization.

Despite of great success, previous studies restricted themselves to a small search space of neural networks. For example, most NAS systems were designed to find an architecture of recurrent or convolutional cell, but the remaining parts of the network are handcrafted \cite{zhong2018block,brock2018smash,elsken2019efficient}. For a larger search space, \citet{zoph2018transferable} optimized the normal cell (i.e., the cell that preserves the dimensionality of the input) and reduction cell (i.e., the cell that reduces the spatial dimension) simultaneously and explored a larger region of the space than the single-cell search.  But it is still rare to see studies on the issue of search space though it is an important factor to NAS. On the other hand, it has been proven that the additional connections between cells help in RNN or Transformer-based models \cite{he2016residual,huang2017dense,wang2018fusion,wang2019dlcl}. These results motivate us to take a step toward the automatic design of inter-cell connections and thus search in a larger space of neural architectures.

\section{Inter-Cell and Intra-Cell NAS}

In this work we use RNNs for description. We choose RNNs because of their effectiveness at preserving past inputs for sequential data processing tasks. Note that although we will restrict ourselves to RNNs for our experiments, the method and discussion here can be applied to other types of models.

\subsection{Problem Statement}

For a sequence of input vectors $\{x_1,...,x_T\}$, an RNN makes a cell on top of every input vector. The RNN cell receives information from previous cells and input vectors. The output at time step $t$ is defined to be:

\begin{equation}
h_t = \pi(\hat{h}_{t-1}, \hat{x}_t)
\end{equation}

\noindent where $\pi(\cdot)$ is the function of the cell. $\hat{h}_{t-1}$ is the representation vector of previous cells, and $\hat{x}_t$ is the representation vector of the inputs up to time step $t$. More formally, we define $\hat{h}_{t-1}$  and $\hat{x}_t$ as functions of cell states and model inputs, like this

\begin{eqnarray}
\hat{h}_{t-1} & = & f(h_{[0,t-1]}; x_{[1,t-1]}) \label{eq:f} \\
\hat{x}_{t} & = & g(x_{[1,t]}; h_{[0,t-1]}) \label{eq:g}
\end{eqnarray}

\noindent where $h_{[0,t-1]} = \{h_{0},...,h_{t-1}\}$ and $x_{[1,t-1]} = \{x_{1},...,x_{t-1}\}$. $f(\cdot)$ models the way that we pass information from previous cells to the next. Likewise, $g(\cdot)$ models the case of input vectors. These functions offer a general method to model connections between cells. For example, one can obtain a vanilla recurrent model by setting $\hat{h}_{t-1} = h_{t-1}$ and $\hat{x}_t = x_t$, while more intra-cell connections can be considered if sophisticated functions are adopted for $f(\cdot)$ and  $g(\cdot)$.

While previous work focuses on searching for the desirable architecture design of $\pi(\cdot)$, we take $f(\cdot)$ and $g(\cdot)$ into account and describe a more general case here. We separate two sub-problems out from NAS for conceptually cleaner description:

\begin{itemize}
\item \textbf{Intra-Cell NAS}. It learns the architecture of a cell (i.e., $\pi(\cdot)$).
\item \textbf{Inter-Cell NAS}. It learns the way of connecting the current cell with previous cells and input vectors (i.e., $f(\cdot)$ and $g(\cdot)$).
\end{itemize}

In the following, we describe the design and implementation of our inter-cell and intra-cell NAS methods.

\begin{figure}[t!]
  \centering
  \tikzstyle{node1} = [rounded corners=3pt,inner sep=4pt,minimum height=6em,minimum width=9em,draw,thick,fill=lightgray!20]
  %\tikzstyle{node2} = [rounded corners=3pt,inner sep=3pt,minimum height=1em,minimum width=1.8em,draw,thick,fill=lightgray!5]
  \tikzstyle{cellnode} = [minimum size=0.4em,inner sep=4pt,circle,draw,thick,fill=lightgray!5]
  \tikzstyle{centernode} = [minimum size=0.2em,inner sep=1pt,circle,thick,fill=red]
  \tikzstyle{node2} = [minimum size=0.9em,inner sep=1pt,circle,thick,fill=white,draw]
  \tikzstyle{standard} = [rounded corners=3pt,thick]
  \centering
    \begin{tikzpicture}
        \node [node1,anchor=west] (h1) at (0,0) {};
        \node [node1,anchor=west] (h2) at ([xshift=1em]h1.east) {};
        
        \node [cellnode,anchor=center] (mul) at ([xshift=0.5em,yshift=4.5em]h1.east) {};
        \node [centernode,anchor=center] (center) at ([xshift=0.5em,yshift=4.5em]h1.east) {};
        
        \draw [->,standard] ([yshift=0.1em]h1.north) -- ([yshift=1.5em]h1.north) -- ([xshift=4.5em,yshift=1.5em]h1.north);
        \draw [->,standard] ([yshift=0.1em]h2.north) -- ([yshift=1.5em]h2.north) -- ([xshift=-4.5em,yshift=1.5em]h2.north);
        \draw [->,thick] ([yshift=0.1em]mul.north) -- ([yshift=0.8em]mul.north);
        \node [anchor=center] (label3) at ([yshift=2em]center.north) {$F(\alpha,\beta)$};
        
        %left
        \node [node2,anchor=south] (c1) at ([xshift=-3em,yshift=0.5em]h1.south) {};
        \node [node2,anchor=south] (c2) at ([xshift=-0.5em,yshift=0.5em]h1.south) {};
        \node [node2,anchor=south] (c3) at ([xshift=3em,yshift=0.5em]h1.south) {};
        \node [anchor=south] (dot1) at ([xshift=1.2em,yshift=0.8em]h1.south) {...};
        \node [anchor=south] (dot2) at ([xshift=1.2em,yshift=-1em]h1.south) {...};
        
        \node [node2,anchor=south,fill=blue!20] (c4) at ([xshift=0.3em,yshift=0.6em]c1.north) {};
        \node [node2,anchor=south,fill=red!20] (c5) at ([xshift=0.5em,yshift=1.4em]c2.north) {};
        \node [node2,anchor=south,fill=green!20] (c6) at ([xshift=0.2em,yshift=1em]c3.north) {};
        
        \node [node2,anchor=south,fill=orange!20] (c7) at ([xshift=-0.5em,yshift=0.6em]c4.north) {};
        %\node [node2,anchor=south,fill=yellow!20] (c8) at ([xshift=4em,yshift=0.8em]c4.north) {};
        \node [node2,anchor=north] (c9) at ([yshift=-0.5em]h1.north) {};
        
        \node [anchor=north] (label1) at ([yshift=-1em]h1.south) {$\alpha$};
        \node [anchor=north] (label11) at ([xshift=3.5em,yshift=-0.2em]h1.north) {$S^{\alpha}$};

        \draw [->,thick] ([yshift=-1.5em]c1.south) -- ([yshift=-0.6em]c1.south);
        \draw [->,thick] ([yshift=-1.5em]c2.south) -- ([yshift=-0.6em]c2.south);
        \draw [->,thick] ([yshift=-1.5em]c3.south) -- ([yshift=-0.6em]c3.south);

        \draw [->,thick] (c1.90) -- (c4.250);
        \draw [->,thick] (c2.100) -- (c4.320);
        \draw [->,thick] (c2.100) -- (c6.220);
        \draw [->,thick] (c3.100) -- (c5.310);
        \draw [->,thick] (c4.120) -- (c7.270);
        \draw [->,thick] (c5.150) -- (c7.330);
        \draw [->,thick] (c6.90) -- (c9.320);
        \draw [->,thick] (c7.60) -- (c9.210);

        % \draw [->,thick] ([yshift=0em]c1.north) -- ([yshift=0em]c4.south);
        % \draw [->,thick] ([yshift=0em]c2.north) -- ([yshift=0em]c4.south);
        % \draw [->,thick] ([yshift=0em]c2.north) -- ([yshift=0em]c6.south);
        % \draw [->,thick] ([yshift=0em]c3.north) -- ([yshift=0em]c5.south);
        % \draw [->,thick] ([yshift=0em]c4.north) -- ([yshift=0em]c7.south);
        % \draw [->,thick] ([yshift=0em]c5.north) -- ([xshift=-0.1em,yshift=0em]c7.south);
        % \draw [->,thick] ([yshift=0em]c6.north) -- ([xshift=0.1em,yshift=0em]c9.south);
        % \draw [->,thick] ([yshift=0em]c7.north) -- ([xshift=-0.1em,yshift=0em]c9.south);
        % %\draw [->,thick] ([yshift=0em]c8.north) -- ([xshift=0.1em,yshift=0em]c9.south);

        %right
        
        \node [node2,anchor=south] (r1) at ([xshift=-3em,yshift=0.5em]h2.south) {};
        \node [node2,anchor=south] (r2) at ([xshift=-0.5em,yshift=0.5em]h2.south) {};
        \node [node2,anchor=south] (r3) at ([xshift=3em,yshift=0.5em]h2.south) {};
        \node [anchor=south] (dot3) at ([xshift=1.2em,yshift=0.8em]h2.south) {...};
        \node [anchor=south] (dot4) at ([xshift=1.2em,yshift=-1em]h2.south) {...};

        \node [node2,anchor=south,fill=orange!20] (r4) at ([xshift=-0.3em,yshift=1.2em]r1.north) {};
        \node [node2,anchor=south,fill=yellow!20] (r5) at ([xshift=-0.1em,yshift=0.6em]r2.north) {};
        \node [node2,anchor=south,fill=blue!20] (r6) at ([xshift=0.3em,yshift=0.6em]r3.north) {};

        \node [node2,anchor=south,fill=red!20] (r7) at ([xshift=5em,yshift=0.2em]r4.north) {};
        \node [node2,anchor=north] (r8) at ([yshift=-0.5em]h2.north) {};
        
        \node [anchor=north] (label2) at ([yshift=-1em]h2.south) {$\beta$};
        \node [anchor=center] (label22) at ([xshift=3.5em,yshift=-0.8em]h2.north) {$S^{\beta}$};
        
        \draw [->,thick] ([yshift=-1.5em]r1.south) -- ([yshift=-0.6em]r1.south);
        \draw [->,thick] ([yshift=-1.5em]r2.south) -- ([yshift=-0.6em]r2.south);
        \draw [->,thick] ([yshift=-1.5em]r3.south) -- ([yshift=-0.6em]r3.south);

        \draw [->,thick] (r1.90) -- (r4.270);
        \draw [->,thick] (r1.80) -- (r5.220);
        \draw [->,thick] (r2.90) -- (r6.220);
        \draw [->,thick] (r3.90) -- (r6.270);
        \draw [->,thick] (r4.90) -- (r8.210);
        \draw [->,thick] (r5.30) -- (r7.230);
        \draw [->,thick] (r6.90) -- (r7.330);
        \draw [->,thick] (r7.140) -- (r8.330);
        % \draw [->,thick] ([yshift=0em]r1.north) -- ([yshift=0em]r4.south);
        % \draw [->,thick] ([yshift=0em]r2.north) -- ([yshift=0em]r5.south);
        % \draw [->,thick] ([yshift=0em]r3.north) -- ([yshift=0em]r6.south);
        
        % \draw [->,thick] ([yshift=0em]r4.north) -- ([xshift=-0.1em,yshift=0em]r7.south);
        % \draw [->,thick] ([yshift=0em]r5.north) -- ([yshift=0em]r7.south);
        % \draw [->,thick] ([yshift=0em]r6.north) -- ([xshift=0.1em,yshift=0em]r7.south);
        
        % \draw [->,thick] ([yshift=0em]r7.north) -- ([yshift=0em]r8.south);

    \end{tikzpicture}

  \caption{Formalizing  intra and inter-cell NAS as learning function $F(\cdot)$.} \label{fig:general-method}
\end{figure}
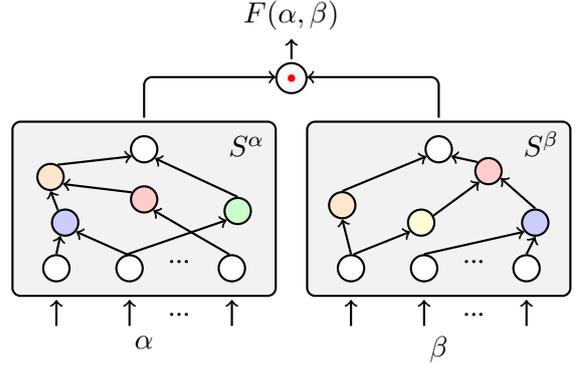

\begin{figure*}[t!]
  \centering
  \tikzstyle{rnnnode1} = [rounded corners=2pt,inner sep=4pt,minimum height=3em,minimum width=3em,draw,thick,fill=blue!10!white]
  \tikzstyle{node1} = [rounded corners=2pt,inner sep=4pt,minimum height=1.2em,minimum width=2em,draw,thick]
  \tikzstyle{node2} = [rounded corners=2pt,inner sep=1pt,minimum height=2em,minimum width=1.5em,draw,thick]
  \tikzstyle{intra} = [rounded corners=2pt,inner sep=4pt,minimum height=6.5em,minimum width=18.5em,draw,thick,fill=blue!5!white]
  \tikzstyle{inter} = [rounded corners=2pt,inner sep=4pt,minimum height=7.5em,minimum width=18.5em,draw,thick,fill=red!5!white]
  \tikzstyle{cellnode} = [minimum size=2em,circle,draw,thick,fill=red!10!white]
  \tikzstyle{label_inter} = [minimum size=0.5em,circle,draw,thick,fill=red!10!white]
  \tikzstyle{label_intra} = [rounded corners=2pt,inner sep=4pt,minimum height=6.5em,minimum width=18.5em,draw,thick,fill=blue!5!white]
  \tikzstyle{standard} = [rounded corners=2pt,thick]
  \centering
    \begin{tikzpicture}
        %left
        \node [rnnnode1,anchor=west] (ht) at (0,0) {};
        \node [cellnode,anchor=east] (circle1) at ([xshift=-4em]ht.west) {};
        \node [cellnode,anchor=north] (circle2) at ([yshift=-5em]ht.south) {};
        %\node [label_inter,anchor=south] (label_inter) at ([xshift=-3em,yshift=3.5em]circle1.north) {};
        %\node [label_inter,anchor=south] (label_inter) at ([xshift=-3em,yshift=3.5em]circle1.north) {};

        %label
        \node [anchor=east] (ht1) at ([xshift=-2em]circle1.west) {$h_{t-1}$};
        \node [anchor=north] (xt) at ([yshift=-1.8em]circle2.south) {$x_{t}$};
        \node [anchor=north] (yt) at ([yshift=4em]ht.north) {$y_{t}$};
        \node [anchor=north] (xhat) at ([xshift=0.8em,yshift=-0.8em]ht.south) {$\hat{x}_{t}$};
        \node [anchor=east] (hhat) at ([xshift=-0.6em,yshift=1em]ht.west) {$\hat{h}_{t-1}$};
        \node [anchor=west] (hhat) at ([xshift=0.5em,yshift=-0.8em]ht.east) {$h_{t}$};

        \node [anchor=north] (x1) at ([xshift=-3em,yshift=-1.2em]circle1.south) {$x_{t-1}$};
        \node [anchor=north] (x2) at ([yshift=-1.2em]circle1.south) {$x_{t-2}$};
        \node [anchor=north] (x3) at ([xshift=3em,yshift=-1.2em]circle1.south) {$x_{t-3}$};

        \node [anchor=east] (h1) at ([xshift=-1.2em,yshift=2.5em]circle2.west) {$h_{t-1}$};
        \node [anchor=east] (h2) at ([xshift=-1.2em]circle2.west) {$h_{t-2}$};
        \node [anchor=east] (h3) at ([xshift=-1.2em,yshift=-2.5em]circle2.west) {$h_{t-3}$};
        \node [anchor=west] (dot1) at ([xshift=2em]ht.east) {$...$};

        \draw [->,thick] ([xshift=0.1em]ht1.east) -- ([xshift=-0.1em]circle1.west);
        \draw [->,thick] ([xshift=0.1em]circle1.east) -- ([xshift=-0.1em]ht.west);
        \draw [->,thick] ([yshift=0.1em]xt.north) -- ([yshift=-0.1em]circle2.south);
        \draw [->,thick] ([yshift=0.1em]circle2.north) -- ([yshift=-0.1em]ht.south);

        \draw [->,thick] ([yshift=0.1em]ht.north) -- ([yshift=-0.1em]yt.south);
        \draw [->,thick] ([xshift=0.1em]ht.east) -- ([xshift=2em]ht.east);

        % \draw [->,thick] ([yshift=0.1em]x1.north) .. controls +(north:0.4) and +(south:0.5) ..(circle1.250);
        % \draw [->,thick] ([yshift=0.1em]x2.north) -- (circle1.270);
        % \draw [->,thick] ([yshift=0.1em]x3.north) .. controls +(north:0.4) and +(south:0.5) ..(circle1.290);

        \draw [->,thick] ([yshift=0.1em]x1.north) -- (circle1.250);
        \draw [->,thick] ([yshift=0.1em]x2.north) -- (circle1.270);
        \draw [->,thick] ([yshift=0.1em]x3.north) -- (circle1.290);

        % \draw [->,thick] ([xshift=0.1em]h1.east) .. controls +(east:0.4) and +(west:0.5) ..(circle2.160);
        % \draw [->,thick] ([xshift=0.1em]h2.east) -- (circle2.180);
        % \draw [->,thick] ([xshift=0.1em]h3.east) .. controls +(east:0.4) and +(west:0.5) ..(circle2.200);

        \draw [->,thick] ([xshift=0.1em]h1.east) -- (circle2.160);
        \draw [->,thick] ([xshift=0.1em]h2.east) -- (circle2.180);
        \draw [->,thick] ([xshift=0.1em]h3.east) -- (circle2.200);

        %right
        \node [intra,anchor=south west,drop shadow] (intra) at ([xshift=7.5em]ht.south east) {};
        \node [inter,anchor=south west,drop shadow] (inter) at ([xshift=7.5em,yshift=-9em]ht.south east) {};

        \draw [->,thick,densely dotted] ([xshift=0.1em,yshift=1em]ht.east) .. controls +(east:1.5) and +(west:1.5) ..([xshift=-0.1em,yshift=2.5em]intra.west);
        \draw [->,thick,densely dotted] (circle2.0) .. controls +(east:1.5) and +(west:2) ..([xshift=-0.1em,yshift=3.5em]inter.west);

        %up
        \node [node1,anchor=west,fill=green!20] (e0) at ([xshift=0.8em]intra.west) {$e_{1}$};
        \node [node1,anchor=west,fill=red!20] (s1) at ([xshift=1em,yshift=1.5em]e0.east) {$s_{1}$};
        \node [node1,anchor=west,fill=red!20] (s2) at ([xshift=3.5em,yshift=-1.5em]e0.east) {$s_{2}$};
        \node [node1,anchor=west,fill=red!20] (s3) at ([xshift=6em,yshift=1.5em]e0.east) {$s_{3}$};

        \node [node2,anchor=west,fill=blue!30] (k1) at ([xshift=12.8em]e0.east) {$s_n$};

        \node [anchor=east] (label0) at ([xshift=-1.5em]e0.west) {$\hat{h}_{t-1}$};
        \node [anchor=north] (label1) at ([xshift=7em,yshift=-2.5em]e0.south) {$\hat{x}_{t}$};
        \node [anchor=west] (label2) at ([xshift=0.2em]intra.east) {$h_{t}$};

        \node [anchor=north] (IntraCell) at ([xshift=-7em,yshift=0.1em]intra.north) {\footnotesize{\textbf{Intra-cell}}};

        \node [anchor=west] (dot2) at ([xshift=7.4em]e0.east) {$...$};

        \draw [->,thick] ([xshift=-0.1em]label0.east) -- ([xshift=-0.1em]e0.west);

        %e1-s1
        \draw [->,standard,densely dotted] ([yshift=0.1em]e0.north) -- ([yshift=0.8em]e0.north) -- ([xshift=2em,yshift=0.8em]e0.north);
        \draw [->,standard,rounded corners=2pt,red] ([xshift=0.4em,yshift=0.1em]e0.north) -- ([xshift=0.4em,yshift=0.4em]e0.north) -- ([xshift=2em,yshift=0.4em]e0.north);
        \draw [->,standard,densely dotted] ([xshift=-0.4em,yshift=0.1em]e0.north) -- ([xshift=-0.4em,yshift=1.2em]e0.north) -- ([xshift=2em,yshift=1.2em]e0.north);

        %s1-s2
        \draw [->,standard,densely dotted] ([yshift=-0.1em]s1.south) -- ([yshift=-2.3em]s1.south) -- ([xshift=1.4em,yshift=-2.3em]s1.south);
        \draw [->,standard,densely dotted] ([xshift=0.4em,yshift=-0.1em]s1.south) -- ([xshift=0.4em,yshift=-1.9em]s1.south) -- ([xshift=1.4em,yshift=-1.9em]s1.south);
        \draw [->,standard,red] ([xshift=-0.4em,yshift=-0.1em]s1.south) -- ([xshift=-0.4em,yshift=-2.7em]s1.south) -- ([xshift=1.4em,yshift=-2.7em]s1.south);

        %s1-s3
        \draw [->,thick,densely dotted] ([xshift=0.1em]s1.east) -- ([xshift=-0.1em]s3.west);
        \draw [->,thick,red] ([xshift=0.1em,yshift=0.4em]s1.east) -- ([xshift=-0.1em,yshift=0.4em]s3.west);
        \draw [->,thick,densely dotted] ([xshift=0.1em,yshift=-0.4em]s1.east) -- ([xshift=-0.1em,yshift=-0.4em]s3.west);

        %s2-s3
        \draw [->,standard,red] ([xshift=0.1em]s2.east) -- ([xshift=1.5em]s2.east) -- ([xshift=1.5em,yshift=2.2em]s2.east);
        \draw [->,standard,densely dotted] ([xshift=0.1em,yshift=-0.4em]s2.east) -- ([xshift=1.9em,yshift=-0.4em]s2.east) -- ([xshift=1.9em,yshift=2.2em]s2.east);
        \draw [->,standard,densely dotted] ([xshift=0.1em,yshift=0.4em]s2.east) -- ([xshift=1.1em,yshift=0.4em]s2.east) -- ([xshift=1.1em,yshift=2.2em]s2.east);

        %avg
        \draw [->,standard] ([xshift=0.1em]s3.east) -- ([xshift=0.8em]s3.east) -- ([xshift=0.8em,yshift=-1.5em]s3.east) -- ([xshift=4.6em,yshift=-1.5em]s3.east);
        \draw [->,standard] ([yshift=0.1em]s1.north) -- ([yshift=0.5em]s1.north) -- ([xshift=7.2em,yshift=0.5em]s1.north) -- ([xshift=7.2em,yshift=-1.5em]s1.north) -- ([xshift=10.65em,yshift=-1.5em]s1.north);
        \draw [->,standard] ([yshift=-0.1em]s2.south) -- ([yshift=-0.5em]s2.south) -- ([xshift=4.3em,yshift=-0.5em]s2.south) -- ([xshift=4.3em,yshift=1.5em]s2.south) -- ([xshift=8.15em,yshift=1.5em]s2.south);

        \node [node2,anchor=west,fill=orange!30] (avg1) at ([xshift=10em]e0.east) {\small{$Avg$}};

        \draw [->,thick] ([xshift=0.1em]k1.east) -- ([xshift=1.8em]k1.east);
        \draw [->,standard] ([xshift=-0.1em]label1.west) -- ([xshift=-6.2em]label1.west) -- ([xshift=-6.2em,yshift=3.2em]label1.west);

        %down
        \node [node1,anchor=west,fill=green!20] (e1) at ([xshift=0.8em,yshift=2em]inter.west) {$e_{1}$};
        \node [node1,anchor=north,fill=green!20] (e2) at ([yshift=-0.8em]e1.south) {$e_{2}$};
        \node [node1,anchor=north,fill=green!20] (e3) at ([yshift=-0.8em]e2.south) {$e_{3}$};

        \node [node1,anchor=west,fill=red!20] (s11) at ([xshift=2.0em]e1.east) {$s_{1}$};
        \node [node1,anchor=west,fill=red!20] (s12) at ([xshift=2.0em]e2.east) {$s_{2}$};
        \node [node1,anchor=west,fill=red!20] (s13) at ([xshift=2.0em]e3.east) {$s_{3}$};
        \node [node1,anchor=west,fill=red!20] (s14) at ([xshift=2.0em]s12.east) {$s_{4}$};

        \node [node2,anchor=west,fill=blue!30] (k2) at ([xshift=4.8em]s14.east) {$s_n$};

        \node [anchor=east] (label1) at ([xshift=-1.5em]e1.west) {$h_{t-1}$};
        \node [anchor=east] (label2) at ([xshift=-1.5em]e2.west) {$h_{t-2}$};
        \node [anchor=east] (label3) at ([xshift=-1.5em]e3.west) {$h_{t-3}$};

        \node [anchor=north] (odot) at ([xshift=0.7em,yshift=4em]k2.south) {$\odot$};

        \node [anchor=north] (InterCell) at ([xshift=-7em,yshift=0.1em]inter.north) {\footnotesize{\textbf{Inter-cell}}};

        \node [anchor=west] (dot3) at ([xshift=-0.5em,yshift=1em]s14.east) {$...$};

        %\draw [->,thick] ([xshift=0.1em]ht.east) -- ([xshift=1.5em]ht.east);

        \foreach \x in {1,2,3}{
            \draw [->,thick] ([xshift=-0.1em]label\x.east) -- ([xshift=-0.1em]e\x.west);
            \draw [->,thick,red] ([xshift=0.1em]e\x.east) -- ([xshift=-0.1em]s1\x.west);
            \draw [->,thick,densely dotted] ([xshift=0.1em,yshift=0.4em]e\x.east) -- ([xshift=-0.1em,yshift=0.4em]s1\x.west);
            \draw [->,thick,densely dotted] ([xshift=0.1em,yshift=-0.4em]e\x.east) -- ([xshift=-0.1em,yshift=-0.4em]s1\x.west);
        }

        \draw [->,thick,densely dotted] ([xshift=0.1em,yshift=0.4em]s12.east) -- ([xshift=-0.1em,yshift=0.4em]s14.west);
        \draw [->,thick,red] ([xshift=0.1em]s12.east) -- ([xshift=-0.1em]s14.west);
        \draw [->,thick,densely dotted] ([xshift=0.1em,yshift=-0.4em]s12.east) -- ([xshift=-0.1em,yshift=-0.4em]s14.west);

        %s1-s4
        \draw [->,thick,densely dotted,standard] ([xshift=0.1em]s11.east) -- ([xshift=3em]s11.east) -- ([xshift=3em,yshift=-1.4em]s11.east);
        \draw [->,thick,densely dotted,standard] ([xshift=0.1em,yshift=0.4em]s11.east) -- ([xshift=3.4em,yshift=0.4em]s11.east) -- ([xshift=3.4em,yshift=-1.4em]s11.east);
        \draw [->,thick,red,standard] ([xshift=0.1em,yshift=-0.4em]s11.east) -- ([xshift=2.6em,yshift=-0.4em]s11.east) -- ([xshift=2.6em,yshift=-1.4em]s11.east);

        %s3-s4
        \draw [->,thick,densely dotted,standard] ([xshift=0.1em]s13.east) -- ([xshift=3em]s13.east) -- ([xshift=3em,yshift=1.4em]s13.east);
        \draw [->,thick,red,standard] ([xshift=0.1em,yshift=-0.4em]s13.east) -- ([xshift=3.4em,yshift=-0.4em]s13.east) -- ([xshift=3.4em,yshift=1.4em]s13.east);
        \draw [->,thick,densely dotted,standard] ([xshift=0.1em,yshift=0.4em]s13.east) -- ([xshift=2.6em,yshift=0.4em]s13.east) -- ([xshift=2.6em,yshift=1.4em]s13.east);

        \draw [->,standard] ([yshift=0.1em]s11.north) -- ([yshift=0.5em]s11.north) -- ([xshift=6em,yshift=0.5em]s11.north) -- ([xshift=6em,yshift=-2.1em]s11.north) -- ([xshift=9.8em,yshift=-2.1em]s11.north);
        \draw [->,standard] ([yshift=-0.1em]s12.south) -- ([yshift=-0.5em]s12.south) -- ([xshift=5.5em,yshift=-0.5em]s12.south) -- ([xshift=5.5em,yshift=0.5em]s12.south) -- ([xshift=9.8em,yshift=0.5em]s12.south);
        \draw [->,standard] ([yshift=-0.1em]s13.south) -- ([yshift=-0.4em]s13.south) -- ([xshift=6em,yshift=-0.4em]s13.south) -- ([xshift=6em,yshift=2.2em]s13.south) -- ([xshift=9.8em,yshift=2.2em]s13.south);
        \draw [->,thick] ([xshift=0.1em,yshift=0.3em]s14.east) -- ([xshift=4.7em,yshift=0.3em]s14.east);
        \draw [->,standard] ([xshift=0.1em]k2.east) -- ([xshift=0.5em]k2.east) -- ([xshift=0.5em,yshift=4.7em]k2.east) -- ([xshift=-7.8em,yshift=4.7em]k2.east);

        \node [node2,anchor=west,fill=orange!30] (avg2) at ([xshift=2em]s14.east) {\small{$Avg$}};
        \node [anchor=north] (xt1) at ([yshift=-2.7em]k2.south) {$x_{t}$};

        \draw [->,thick] ([yshift=-.1em]xt1.north) -- ([yshift=-0.1em]k2.south);

    \end{tikzpicture}

  \caption{An example of intra-cell and inter-cell NAS in RNN models.} \label{fig:inter-intra-search}
\end{figure*}
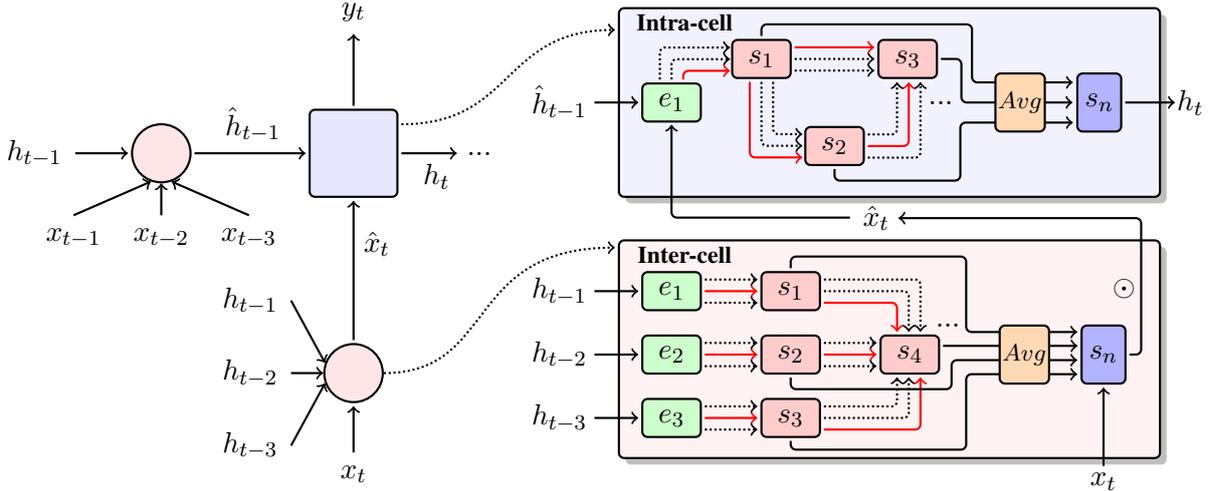

\subsection{Differentiable Architecture Search}

For search algorithms, we follow the method of differentiable architecture search (DARTS). It is gradient-based and runs orders of magnitude faster than earlier methods \cite{zoph2018transferable,real2019regularized}. DARTS represents networks as a directed acyclic graph (DAG) and search for the appropriate architecture on it. For a DAG, the edge $o^{i,j}(\cdot)$ between node pair $(i, j)$ performs an operation to transform the input (i.e., tail) to the output (i.e., head). Like \citet{liu2019darts}'s method and others, we choose operations from a list of activation functions, e.g., sigmoid, identity and etc\footnote{We also consider a special activation function ``drop'' that unlinks two nodes.}. A node represents the intermediate states of the networks.  For node $i$, it weights vectors from all predecessor nodes ($j < i$) and simply sums over them. Let $s_i$ be the state of node $i$. We define $s_i$ to be:

\begin{equation}
s_i = \sum_{j<i} \sum_k \theta_k^{i,j}  \cdot  o_k^{i,j}(s_j \cdot W_j) \label{eq:darts-state}
\end{equation}

\noindent where $W_j$ is the parameter matrix of the linear transformation, and $\theta_k^{i,j}$  is the weight indicating the importance of $o_k^{i,j}(\cdot)$. Here the subscript $k$ means the operation index. $\theta_k^{i,j}$ is obtained by softmax normalization over edges between nodes $i$ and $j$: $\theta_k^{i,j} = \exp(w_{k}^{i,j}) / \sum_{k'}\exp(w_{k'}^{i,j})$. In this way, the induction of discrete networks is reduced to learning continuous variables $\{\theta_k^{i,j} \}$ at the end of the search process. This enables the use of efficient gradient descent methods. Such a model encodes an exponentially large number of networks in a graph, and the optimal architecture is generated by selecting the edges with the largest weights.

The common approach to DARTS constraints the output of the generated network to be the last node that averages the outputs of all preceding nodes. Let $s_n$ be the last node of the network. We have

\begin{equation}
s_n = \frac{1}{n-1}\sum_{i=1}^{n-1} s_i
\end{equation}

Given the input vectors, the network found by DARTS generates the result at the final node $s_n$. Here we present a method to fit this model into intra and inter-cell NAS. We re-formalize the function for which we find good architectures as $F(\alpha;\beta)$. $\alpha$ and $\beta$ are two groups of the input vectors. We create DAGs on them individually. This gives us two DAGs with $s^{\alpha}$ and $s^{\beta}$ as the last nodes. Then, we make the final output by a Hadamard product of $s^{\alpha}$ and $s^{\beta}$, like this,

\begin{equation}
F(\alpha;\beta) = s^{\alpha} \odot s^{\beta}
\end{equation}

See Figure \ref{fig:general-method} for the network of an example $F(\alpha;\beta)$. This method transforms the NAS problem into two learning tasks.  The design of two separate networks allows the model to group related inputs together, rather than putting everything into a ``magic'' system of NAS. For example, for the inter-cell function $f(\cdot)$, it is natural to learn the pre-cell connection from $h_{[0,t-1]}$, and learn the impact of the model inputs from $x_{[1,t-1]}$. It is worth noting that the Hadamard product of $s^{\alpha}$ and $s^{\beta}$ is doing something very similar to the gating mechanism which has been widely used in NLP \cite{dauphin2017gate,bradbury2017qrnn,gehring2017convolutional}. For example, one can learn $s^{\beta}$ as a gate and control how much $s^{\alpha}$ is used for final output. Table \ref{tab:alpha-beta} gives the design of $\alpha$ and $\beta$ for the functions used in this work.

Another note on $F(\alpha;\beta)$. The grouping reduces a big problem into two cheap tasks. It is particularly important for building affordable NAS systems because computational cost increases exponentially as more input nodes are involved. Our method instead has a linear time complexity if we adopt a reasonable constraint on group size, leading to a possibility of exploring a much larger space during the architecture search process.

\begin{table}
\centering
\begin{tabular}{l| ll}
Function  & $\alpha$ & $\beta$ \\
\hline
$\pi(\cdot)$ & $\{\hat{h}_{t-1}, \hat{x}_t\}$  & 1 \\
$f(\cdot)$ & $h_{[0,t-1]}$   & $x_{[1,t-1]}$  \\
$g(\cdot)$ & $x_{[1,t]}$  & $h_{[0,t-1]}$  \\
\end{tabular}
\caption{$\alpha$ and $\beta$ for different functions}
\label{tab:alpha-beta}
\end{table}

\subsection{The Intra-Cell Search Space} \label{sec:in-cell}

The search of intra-cell architectures is trivial. Since $\beta = 1$ and $s^{\beta}=1$ (see Table \ref{tab:alpha-beta}), we are basically performing NAS on a single group of input vectors $\hat{h}_{t-1}$ and $\hat{x}_t$. We follow \citet{liu2019darts}'s work and force the input of networks to be a single layer network of $\hat{h}_{t-1}$ and $\hat{x}_t$. This can be described as

\begin{equation}
\setlength{\abovedisplayskip}{3pt}
\label{eqn:cal-e1}
e_1 =  \textrm{tanh}(\hat h_{t-1} \cdot W^{(h)} + \hat x_t \cdot W^{(x)} )
\end{equation}

\noindent where $W^{(h)}$ and $W^{(x)}$ are parameters of the transformation, and $\textrm{tanh}$ is the non-linear transformation. $e_1$ is the input node of the graph. See Figure \ref{fig:inter-intra-search} for intra-cell NAS of an RNN models.

\subsection{The Inter-Cell Search Space} \label{sec:inter-cell}

To learn $\hat h_{t-1}$ and $\hat x_t$, we can run the DARTS system as described above. However, Eqs. (\ref{eq:f}-\ref{eq:g}) define a model with a varying number of parameters for different time steps, in which our architecture search method is not straightforwardly applicable. Apart from this, a long sequence of RNN cells makes the search intractable.

\begin{figure}
\begin{tabular}{l}
\hline
\hspace{0em}\textbf{Function} \textsc{JointLearn} ($rounds$, $w$, $W$) \\
\hline
1: \hspace{0em} \textbf{for} $i$ in range(1, $rounds$) \textbf{do} \\
2: \hspace{1em} \textbf{while} intra-cell $model$ not converged \textbf{do} \\
3: \hspace{2em} Update intra-cell $w^{(intra)}$ and $W$ \\
4: \hspace{1em} \textbf{while} inter-cell $model$ not converged \textbf{do} \\
5: \hspace{2em} Update inter-cell $w^{(inter)}$ and $W$ \\
6: \hspace{0em} Derive $architecture$ based on $w$ \\
7: \hspace{0em} \textbf{return} $architecture$\\
\hline
\end{tabular}
\caption{Joint search of intra-cell and inter-cell architectures. $w$ = edge weights, and $W$ = model parameters.}
\label{fig:joint-learning}
\end{figure}

For a simplified model, we re-define $f(\cdot)$ and $g(\cdot)$ as:

\begin{eqnarray}
f(h_{[0,t-1]};x_{[1,t-1]}) & \hspace{-0.4em} = & \hspace{-0.6em} f'(h_{t-1}; x_{[t-m,t-1]}) \label{eq:fspy} \\
g(x_{[1,t]};h_{[0,t-1]}) & \hspace{-0.4em} = & \hspace{-0.6em} g'(x_{t}; h_{[t-m,t-1]}) \label{eq:gspy}
\end{eqnarray}

\noindent where $m$ is a hyper-parameter that determines how much history is considered. Eq. (\ref{eq:fspy}) indicates a model that learns a network on $x_{[t-m,t-1]}$ (i.e., $\beta=x_{[t-m,t-1]}$). Then, the output of the learned network (i.e., $s^\beta$) is used as a gate to control the information that we pass from the previous cell to the current cell (i.e., $\alpha=\{h_{t-1}\}$). Likewise, Eq. (\ref{eq:gspy}) defines a gate on $h_{[t-m,t-1]}$ and controls the information flow from $x_t$ to the current cell.

Learning $f'(\cdot)$ and $g'(\cdot)$ fits our method well due to the fixed number of input vectors. Note that $f'(\cdot)$ has $m$ input vectors $x_{[t-m,t-1]}$ for learning the gate network. Unlike what we do in intra-cell NAS, we do not concatenate them into a single input vector. Instead, we create a node for every input vector, that is, the input vector $e_i=x_{t-i}$ links with node $s_i$. We restrict $s_i$ to only receive inputs from $e_i$ for better processing of each input. This can be seen as a pruned network for the model described in Eq. (\ref{eq:darts-state}). See Figure \ref{fig:inter-intra-search} for an illustration of inter-cell NAS.

\section{Joint Learning for Architecture Search}

Our model is flexible. For architecture search, we can run intra-cell NAS, or inter-cell NAS, or both of them as needed. However, we found that simply joining intra-cell and inter-cell architectures might not be desirable because both methods were restricted to a particular region of the search space, and the simple combination of them could not guarantee the global optimum.

This necessitates the inclusion of interactions between intra-cell and inter-cell architectures into the search process. Generally, the optimal inter-cell architecture depends on the intra-cell architecture used in search, and vice versa. A simple method that considers this issue is to learn two models in a joint manner. Here, we design a joint search method to make use of the interaction between intra-cell NAS and inter-cell NAS. Figure \ref{fig:joint-learning} shows the algorithm. It runs for a number of rounds. In each round, we first learn an optimal intra-cell architecture by fixing the inter-cell architecture, and then learn a new inter-cell architecture by fixing the optimal intra-cell architecture that we find just now.

Obviously, a single run of intra-cell (or inter-cell) NAS is a special case of our joint search method. For example, one can turn off the inter-cell NAS part (lines 4-5 in Figure \ref{fig:joint-learning}) and learn intra-cell architectures solely. In a sense, the joint NAS method extends the search space of individual intra-cell (or inter-cell) NAS. Both intra-cell and inter-cell NAS shift to a new region of the parameter space in a new round. This implicitly explores a larger number of underlying models. As shown in our experiments, joint NAS learns intra-cell architectures unlike those of the individual intra-cell NAS, which leads to better performance in language modeling and other tasks.

\section{Experiments}
We experimented with our ESS method on Penn Treebank and WikiText language modeling tasks and applied the learned architecture to NER and chunking tasks to test its transferability.

\subsection{Experimental Setup}

For language modeling task, the monolingual and evaluation data came from two sources.

\begin{itemize}
\item Penn Treebank (PTB). We followed the standard preprocessed version of PTB \cite{mikolov2010rnnlm}. It consisted of 929k training words, 73k validation words and 82k test words. The vocabulary size was set to 10k.
\item WikiText-103 (WT-103). We also used WikiText-103 \cite{merity2017revisiting} data to search for a more universal architecture for NLP tasks. This dataset contained a larger training set of 103 million words and 0.2 million words in the validation and test sets.
\end{itemize}

\begin{table*}[]
    \setlength{\tabcolsep}{1pt}
      %\small{
      \centering
      \begin{tabular}{l|c|c|c|c|c|c|c}
        \hline
        \multicolumn{1}{c|}{\multirow{2}{*}{\centering Dataset}} & \multicolumn{1}{c|}{\multirow{2}{*}{\centering Method}} & \multicolumn{2}{c|}{Search Space} & \multicolumn{1}{c|}{\multirow{2}{*}{\centering Params}} & \multicolumn{2}{c|}{Perplexity} & \multicolumn{1}{c}{\multirow{1}{*}{\centering Search Cost}} \\
        \cline{3-4} \cline{6-8}
  
        \multicolumn{1}{c|}{} & \multicolumn{1}{c|}{} & \multicolumn{1}{c|}{intra-cell} & \multicolumn{1}{c|}{inter-cell} & \multicolumn{1}{c|}{} & \multicolumn{1}{c|}{valid} & \multicolumn{1}{c|}{test} & \multicolumn{1}{c}{(GPU days)} \\
  
        \hline
        \multicolumn{1}{c|}{\multirow{6}{*}{\centering PTB}} 
        & \multirow{1}{*}{AWD-LSTM \cite{merity2018awdlstm}}  & - & - & 24M  & 61.2 & 58.8  & - \\
        & \multirow{1}{*}{Transformer-XL \cite{transformerxl}}  & - & - & 24M  & 56.7 & 54.5  & - \\
        & \multirow{1}{*}{Mogrifier LSTM \cite{melis2019mogrifier}}  & - & - & 23M  & \bf 51.4 & \bf 50.1  & - \\
        \cline{2-8}
        & \multirow{1}{*}{ENAS \cite{pham2018parameter}}  & \cmark & - & 24M & 60.8 & 58.6  & 0.50 \\
        & \multirow{1}{*}{RS \cite{liam2019random}} & \cmark & - & 23M & 57.8 & 55.5  & 2 \\
        & \multirow{1}{*}{DARTS$^\dagger$}  & \cmark & - & 23M & 55.2 & 53.0 & 0.25 \\
        & \multirow{1}{*}{ESS}  & - & \cmark & 23M & 54.1 & 52.3 & 0.5 \\
        & \multirow{1}{*}{ESS}  & \cmark & \cmark & 23M & \bf 47.9 & \bf 45.6 & 0.5 \\
        \hline
        \multicolumn{1}{c|}{\multirow{5}{*}{\centering WT-103}} 
        & \multirow{1}{*}{QRNN \cite{qrnnwiki}}  & - & - & 151M  & 32.0 & 33.0  & - \\
        & \multirow{1}{*}{Hebbian + Cache \cite{hebbianwiki}}  & - & - & -  & 29.9 & 29.7  & - \\
        & \multirow{1}{*}{Transformer-XL \cite{transformerxl}}  & - & - & 151M  & \bf 23.1 & \bf 24.0  & - \\
        \cline{2-8}
        & \multirow{1}{*}{DARTS$^\dagger$}  & \cmark & - & 151M  & 31.4 & 31.6  & 1 \\
        & \multirow{1}{*}{ESS}  & \cmark & \cmark & 156M  & \bf 28.8 & \bf 29.2  & 1.5 \\
        \hline
      \end{tabular}
      \caption{ Comparison of language modeling methods on PTB and WikiText-103 tasks (lower perplexity is better). $^\dagger$Obtained by training the corresponding architecture using our setup.}
      \label{tab:main-result}
      %}
\end{table*}

NER and chunking tasks were also used to test the transferability of the pre-learned architecture. We transferred the intra and inter-cell networks learned on WikiText-103 to the CoNLL-2003 (English), the WNUT-2017 NER tasks and the CoNLL-2000 tasks. The CoNLL-2003 task focused on the newswire text, while the WNUT-2017 contained a wider range of English text which is more difficult to model. 

Our ESS method consisted of two components, including recurrent neural architecture search and architecture evaluation. During the search process, we ran our ESS method to search for the intra-cell and inter-cell architectures jointly. In the second stage, the learned architecture was trained and evaluated on the test dataset.

For architecture search on language modeling tasks, we applied 5 activation functions as the candidate operations, including drop, identity, sigmoid, tanh and relu. On the PTB modeling task, 8 nodes were equipped in the recurrent cell. For the inter-cell architecture, it received 3 input vectors from the previous cells and consisted of the same number of the intermediate nodes. By default, we trained our ESS models for 50 rounds. We set $batch=256$ and used 300 hidden units for the intra-cell model. The learning rate was set as $3 \times 10^{-3}$ for the intra-cell architecture and $1 \times 10^{-3}$ for the inter-cell architecture. The BPTT \cite{werbos1990backpropagation} length was 35. For the search process on WikiText-103, we developed a more complex model to encode the representation. There were 12 nodes in each cell and 5 nodes in the inter-cell networks. The batch size was 128 and the number of hidden units was 300 which was the same with that on the PTB task. We set the intra-cell and inter-cell learning rate to $1 \times 10^{-3}$ and $1 \times 10^{-4}$. A larger window size ($=70$) for BPTT was applied for the WikiText-103. All experiments were run on a single NVIDIA 1080Ti.

After the search process, we trained the learned architectures on the same data. To make it comparable with previous work, we copied the setup in \citet{merity2018analysis}. For PTB, the size of hidden layers was set as 850 and the training epoch was 3,000. While for the WikiText-103, we enlarged the number of hidden units to 2,500 and trained the model for 30 epochs. Additionally, we transferred the learned architecture to NER and chunking tasks with the setting in \citet{Flair}. We only modified the batch size to 24 and hidden size to 512.

\subsection{Results}

\subsubsection{Language Modeling tasks}

Here we report the perplexity scores, number of parameters and search cost on the PTB and WikiText-103 datasets (Table \ref{tab:main-result}). First of all, the joint ESS method improves the performance on language modeling tasks significantly. Moreover, it does not introduce many parameters. Our ESS method achieves state-of-the-art result on the PTB task. It outperforms the manually designed Mogrifier-LSTM by 4.5 perplexity scores on the test set. On the WikiText task, it still yields a +2.4 perplexity scores improvement over the strong NAS baseline (DARTS) method. These results indicate that ESS is robust and can learn better architectures by enlarging the scope of search space.

\begin{figure}[t]
\centering
\begin{tabular}{p{9.6cm}p{9.6cm}}
\begin{tikzpicture}{baseline}
\footnotesize{
\begin{axis}[
    width=.45\textwidth,
    height=.22\textwidth,
    legend style={at={(0.98,0.35)},anchor=north east,font=\tiny,legend columns=1,inner sep=2pt},
    xlabel={\footnotesize{Number of nodes (intra/inter)}},ylabel={\footnotesize{Perplexity}},
    ylabel style={yshift=-0.8em},xlabel style={yshift=0.8em},
    ymin=59.0,ymax=68.0, ytick={59.5, 63.5, 67.5},
    xmin=0.5,xmax=10.5,xtick={1, 2, 3, 4, 5, 6, 7, 8, 9, 10},xticklabels={10/1, 9/2, 8/3, 7/4, 6/5, 5/6, 4/7, 3/8, 2/9, 1/10},xticklabel style={font=\scriptsize},yticklabel style={font=\scriptsize}
]
\addplot[red,mark=otimes*,line width=0.5pt] coordinates {(1,62.78) (2,62.35) (3,60.89) (4,61.41) (5,62.72) (6,62.60) (7,62.19) (8,65.16) (9,66.35) (10,66.62)};
\addlegendentry{NAS}	
\end{axis}
}
\end{tikzpicture}
\\
\end{tabular}
\caption{Perplexity on the validation data (PTB) vs. number of nodes in intra and inter-cell. }
\label{fig:node-number}
\end{figure}
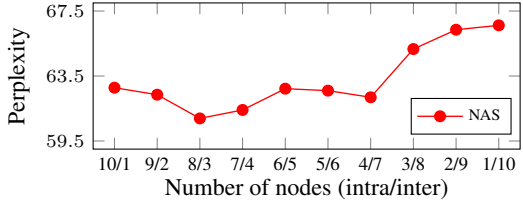

Also, we find that searching for the appropriate connections among cells plays a more important role in improving the model performance. We observe that the intra-cell NAS (DARTS) system underperforms the inter-cell counterpart with the same number of parameters. It is because the well-designed intra-cell architectures (e.g., Mogrifier-LSTM) are actually competitive with the NAS structures. However, the fragile connections among different cells greatly restrict the representation space. The additional inter-cell connections are able to encode much richer context.

Nevertheless, our ESS method does not defeat the manual designed Transformer-XL model on the WikiText-103 dataset, even though ESS works better than other RNN-based NAS methods. This is partially due to the better ability of Transformer-XL to capture the language representation. Note that RNNs are not good at modeling the long-distance dependence even if more history states are considered. It is a good try to apply ESS to Transformer but this is out of the scope of this work.

\subsubsection{Sensitivity Analysis}

To modulate the complexity of the intra and inter-cell, we study the system behaviors under different numbers of intermediate nodes (Figure \ref{fig:node-number}). Fixing the number of model parameters, we compare these systems under different numbers of the intra and inter-cell nodes. Due to the limited space, we show the result on the PTB in the following sensitivity analysis. We observe that an appropriate choice of node number (8 nodes for intra-cell and 3 nodes for inter-cell) brings a consistent improvement. More interestingly, we find that too many nodes for inter-cell architecture do not improve the model representation ability. This is reasonable because more inter-cell nodes refer to considering more history in our system. But for language modeling, the current state is more likely to be relevant to most recent words. Too many inputs to the gate networks raise difficulties in modeling.

\begin{figure}[t]
\centering
\begin{tabular}{p{3.6cm}p{3.6cm}}
\begin{tikzpicture}{baseline}
\footnotesize{
\begin{axis}[
    width=.25\textwidth,
    height=.22\textwidth,
    legend style={at={(0.98,0.97)},anchor=north east,font=\tiny,legend columns=1,inner sep=2pt},
    ylabel style={yshift=-1.2em},xlabel style={yshift=0.8em,align=center},
    xlabel={\footnotesize{\# of Training Steps}},ylabel={\footnotesize{Perplexity}},
    ymin=370,ymax=880, ytick={400, 550, 700, 850},
    xmin=-5,xmax=55,xtick={5, 20, 35, 50}, xticklabels={0.5K, 2K, 3.5K, 5K},xticklabel style={font=\scriptsize},yticklabel style={font=\scriptsize}
]

\addplot[blue,mark=triangle*,line width=0.5pt] coordinates {(5,821.2) (10,651.5) (15,490.8) (20,427.1) (25,422.7) (30,420.4) (35,419.2) (40,418.4) (45,418.0) (50,417.8)};
\addlegendentry{joint}
\addplot[red,mark=otimes*,line width=0.5pt] coordinates { (5,621.6) (10,467.8) (15,465.1) (20,462.6) (25,460.5) (30,459.2) (35,458.5) (40,458.1) (45,457.9) (50,457.8)};
\addlegendentry{intra}
\end{axis}
}
\end{tikzpicture}
&
\begin{tikzpicture}{baseline}
\footnotesize{
\begin{axis}[
    width=.25\textwidth,
    height=.22\textwidth,
    legend style={at={(0.99,0.97)},anchor=north east,font=\tiny,legend columns=2,inner sep=0.5pt},
    ylabel style={yshift=-1.2em},xlabel style={yshift=0.8em, align=center},
    xlabel={\footnotesize{\# of Training Steps}},ylabel={\footnotesize{MAD}},
    ymin=-0.1,ymax=0.7, ytick={0.00, 0.15, 0.30, 0.45, 0.60}, yticklabels={0.00, 0.15, 0.30, 0.45, 0.60}, 
    xmin=-5,xmax=55,xtick={5, 20, 35, 50}, xticklabels={0.5K, 2K, 3.5K, 5K},xticklabel style={font=\scriptsize},yticklabel style={font=\scriptsize}
]
\addplot[blue,mark=triangle*,line width=0.5pt] coordinates {(0.1, 0.000208333) (5,0.016727778) (10,0.062565972) (15,0.138601389) (20,0.215770139) (25,0.279773611) (30,0.324876389) (35,0.356000694) (40,0.376114583) (45,0.389600694) (50,0.398265972)};
\addlegendentry{intra}
\addplot[red,mark=otimes*,line width=0.5pt] coordinates {(0.1, 0.000116667) (5,0.096116667) (10,0.188883333) (15,0.26195) (20,0.315233333) (25,0.354516667) (30,0.38295) (35,0.404516667) (40,0.420691667) (45,0.433408333) (50,0.443291667)};
\addlegendentry{inter}
\end{axis}
}
\end{tikzpicture}
\\
\end{tabular}
\caption{Perplexity on the validation data (PTB) and Mean Absolute Deviation (MAD) between edge weights and uniform distribution vs. number of training steps.}
\label{fig:convergence}
\end{figure}
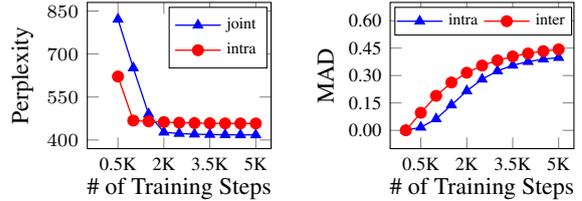

\begin{table}[t]
\setlength{\tabcolsep}{2.0pt}
  %\small
  \centering
  \begin{tabular}{lcc|lcc}
      \hline
      Word & Count & $\Delta$loss & Word & Count & $\Delta$loss\\
      \hline
      mcmoran & 11 & -0.74 & the & 59421 & -0.009 \\
      cie. & 9 & -0.66 & \textless unk \textgreater  & 53299 & -0.004 \\
      mall & 13 & -0.65 & \textless eos \textgreater & 49199 & -0.010 \\
      missile & 23 & -0.55 & N & 37607 & -0.008 \\
      siemens & 12 & -0.51 & of & 28427 & -0.008 \\
      baldwin & 9 & -0.51 & to & 27430 & -0.004\\
      nfl & 21 & -0.49 & a & 24755 & -0.013\\
      prime-time & 17 & -0.47 & in & 21032 & -0.015  \\

      \hline
  \end{tabular}
  \caption{
  Difference in word loss (normalized by word counts) on validation data when searching intra and inter-cell jointly. The left column contains the words with eight best improvements (larger absolute value of $\Delta$loss) and right column presents the most frequent words in the validation data.
  }
  \label{table:loss-word}
\end{table}

\begin{figure*}
     \centering
     \includegraphics[height=4.2cm]{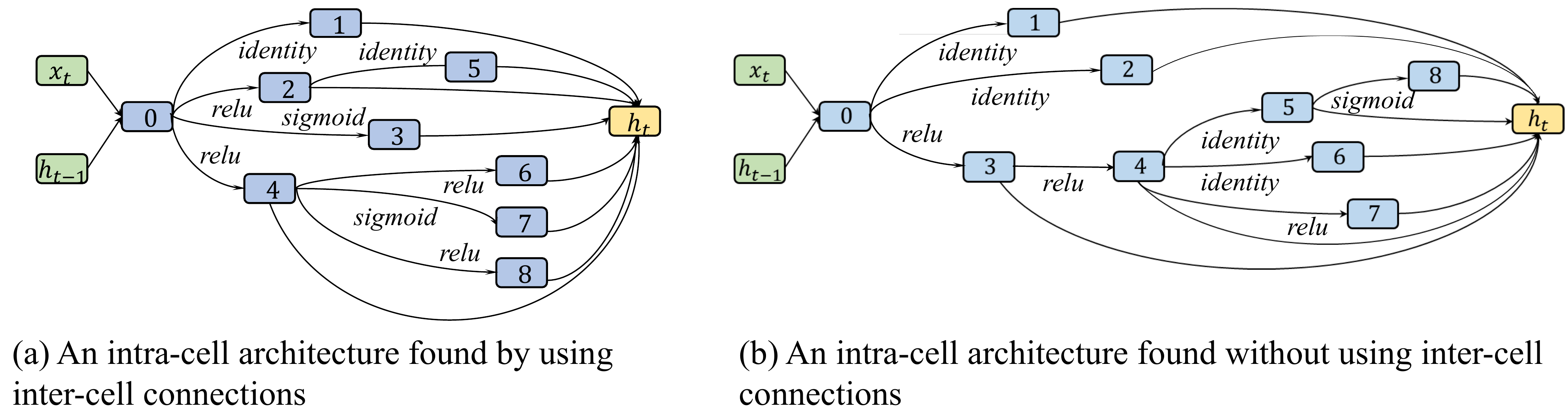}
     \caption{Comparison of intra-cell architectures found by using and not using additional inter-cell connections} \label{fig:intra-cell}
\end{figure*}

We observe that our ESS method leads to a model that is easier to train. The left part in Figure \ref{fig:convergence} plots the validation perplexity at different training steps. The loss curve of joint ESS significantly goes down as the training proceeds. More interestingly, our joint learning method makes the model achieve a lower perplexity than the intra-cell NAS system. This indicates better networks can be obtained in the search process. Additionally, the convergence can be observed from the right part in Figure \ref{fig:convergence}. Here we apply Mean Absolute Deviation (MAD) to define the distance between edge weights and initial uniform distribution. It is obvious that both the intra and inter-cell architectures change little at the final searching steps.

In order to figure out the advantage of inter-cell connections, we detail the model contribution on each word on the validation data. Specifically, we compute the difference in word loss function (i.e., log perplexity) between methods with and without inter-cell NAS. The words with eight best improvements are shown in the left column of Table \ref{table:loss-word}. We observe that the rare words in the training set obtain more significant improvements. In contrast, the most frequent words lead to very modest decrease in loss (right column of Table \ref{table:loss-word}). This is because the connections between multiple cells enable learning rare word representations from more histories. While for common words, they can obtain this information from rich contexts. More inputs from previous cells do not bring much useful information.

Additionally, we visualize the learned intra-cell architecture in Figure \ref{fig:intra-cell}(a). The networks are jointly learned with the inter-cell architecture. Compared with the results of intra-cell NAS (Figure \ref{fig:intra-cell}(b)), the learned network is more shallow. The inter-cell architectures have deeper networks. This in turn reduces the need for intra-cell capacity. Thus a very deep intra-cell architecture might not be necessary if we learn the whole model jointly.

\begin{table}[t]
    \small
    \centering
    \begin{tabular}{lccc}
    \hline 
        Models & F1 \\ \hline
        LSTM-CRF \cite{LSTM-CRF}             &     90.94   \\
        LSTM-CRF + ELMo \cite{ELMo}      &     92.22   \\
        LSTM-CRF + Flair  \cite{Flair}    &     93.18   \\
        GCDT + $\textrm{BERT}_{\textrm{LARGE}}$ \cite{GCDT} &     93.47   \\
        CNN Large + ELMo \cite{CNN-Finetune} &     \bf 93.50   \\
        \hline
        DARTS + Flair   \cite{jiang2019idarts}      &     93.13   \\
        I-DARTS + Flair \cite{jiang2019idarts}      &     93.47   \\
        ESS            & 91.78   \\
        ESS + Flair    & \bf 93.62   \\
        \hline
    \end{tabular}
    \caption{
    F1 scores on CoNLL-2003 NER task. 
 Bi-LSTM
    }
    \label{table:ner-conll}
\end{table}

\begin{table}[t]
    \small
    \centering
    \begin{tabular}{lccc}
    \hline 
        Models & F1 \\ \hline
        Cross-BiLSTM-CNN  \cite{LSTM-CRF+Multitask}   &   45.55   \\
        Flair   \cite{Flair}  &   \bf 50.20   \\
        \hline
        DARTS + Flair$^\dagger$       &   50.34   \\
        ESS          &   48.85   \\
        ESS + Flair  &   \bf 52.18   \\
        \hline
    \end{tabular}
    \caption{
    F1 scores on WNUT-2017 NER task. $^\dagger$Obtained by training the corresponding architecture using our setup.
    }
    \label{table:ner-wnut}
\end{table}

\begin{table}[t]
    \small
    \centering
    \begin{tabular}{lccc}
        \hline 
        Models & F1 \\ \hline
        NCRF++ \cite{NCRF++} & 95.06 \\
        BiLSTM-CRF + IntNet  \cite{IntNet}   &   95.29   \\
        Flair   \cite{Flair}   &   96.72   \\
        GCDT + $\textrm{BERT}_{\textrm{LARGE}}$ \cite{GCDT}  &   \bf 97.30   \\
        \hline
        DARTS + Flair$^\dagger$ & 96.59 \\
        ESS        & 95.51 \\
        ESS + Flair & \bf 97.22 \\
        \hline
    \end{tabular}
    \caption{
        F1 scores on CoNLL-2000 chunking task. $^\dagger$Obtained by training the corresponding architecture using our setup.
        }
    \label{table:chunk}
\end{table}

\subsubsection{Transferring to Other Tasks}

After architecture search, we test the transferability of the learned architecture. In order to apply the model to other tasks, we directly use the architecture searched on WikiText-103 and train the parameters with the in-domain data. In our experiments, we adapt the model to CoNLL-2003,  WNUT-2017 NER tasks and CoNLL-2000 chunking task. 

For the two NER tasks, it achieves new state-of-the-art F1 scores (Table \ref{table:ner-conll} and Table \ref{table:ner-wnut}). ELMo, Flair and $\textrm{BERT}_{\textrm{LARGE}}$ refer to the pre-trained language models. We  apply these word embeddings to the learned architecture during model training process. For the chunking task, the learned architecture also shows greater performance than other NAS methods (Table \ref{table:chunk}). Moreover, we find that our pre-learned neural networks yield bigger improvements on the WNUT-2017 task. The difference of the two NER tasks lies in that the WNUT-2017 task is a long-tail emerging entities recognition task. It focuses on identifying unusual, previously-unseen entities in the context of emerging discussions. As we discuss in the previous part of the section, the additional inter-cell NAS is good at learning the representations of rare words. Therefore, it makes sense to have a bigger improvement on WNUT-2017.

\section{Conclusions}
We have proposed the Extended Search Space (ESS) method of NAS. It learns intra-cell and inter-cell architectures simultaneously. Moreover, we present a general model of differentiable architecture search to handle the arbitrary search space. Meanwhile, the high-level and low-level sub-networks can be learned in a joint fashion. Experiments on two language modeling tasks show that ESS yields improvements of 4.5 and 2.4 perplexity scores over a strong RNN-based baseline. More interestingly, it is observed that transferring the pre-learned architectures to other tasks also obtains a promising performance improvement.

\section*{Acknowledgments}
This work was supported in part by the National Science Foundation of China (Nos. 61876035 and 61732005), the National Key R\&D Program of China (No. 2019QY1801) and the Opening Project of Beijing Key Laboratory of Internet Culture and Digital Dissemination Research. The authors would like to thank anonymous reviewers for their comments.

\bibliography{acl2020}
\bibliographystyle{acl_natbib}

\end{document}